\documentclass{ecai}
\usepackage{graphicx}
\usepackage{latexsym}
\usepackage{booktabs}
\usepackage{amssymb}
\usepackage{amsmath}
\usepackage{adjustbox}

\newcommand{\modelname}{VADES}

\begin{document}

\begin{frontmatter}

\title{Capturing Style in Author and Document Representation}

\author[A]{\fnms{Enzo}~\snm{Terreau}}
\author[B]{\fnms{Antoine}~\snm{Gourru}\thanks{Corresponding Author. Email: antoine.gourru@univ-st-etienne.fr.}}
\author[B]{\fnms{Julien}~\snm{Velcin}}

\address[A]{Université de Lyon, Lyon 2, ERIC UR3083}
\address[B]{Laboratoire Hubert Curien, UMR CNRS 5516, Saint-Etienne, France}

\begin{abstract}
A wide range of Deep Natural Language Processing (NLP) models integrates continuous and low dimensional representations of words and documents. Surprisingly, very few models study representation learning for authors. These representations can be used for many NLP tasks, such as author identification and classification, or in recommendation systems. A strong limitation of existing works is that they do not explicitly capture writing style, making them hardly applicable to literary data. We therefore propose a new architecture based on Variational Information Bottleneck (VIB) that learns embeddings for both authors and documents with a stylistic constraint. Our model fine-tunes a pre-trained document encoder. We stimulate the detection of writing style by adding predefined stylistic features making the representation axis interpretable with respect to writing style indicators.  
We evaluate our method on three datasets: a literary corpus extracted from the Gutenberg Project, the Blog Authorship Corpus and IMDb62, for which we show that it matches or outperforms strong/recent baselines in authorship attribution while capturing much more accurately the authors stylistic aspects.
\end{abstract}

\end{frontmatter}

\section{Introduction}

Deep models for Natural Language Processing are usually based on Transformers, and they rely on latent intermediate representations.
These representations are usually built in a self-supervised manner on a language modeling task, such as Masked Language Modeling (MLM) \cite{devlin2019bert} or auto-regressive training \cite{GPT}.
They constitute a good feature space to solve downstream tasks, for example classification or generation, even though some of those tasks are still difficult to handle with prompt-based generative models like ChatGPT \cite{qin2023chatgpt}.
Additionally, some efforts have been made to benefit from large pretrained model to represent documents \cite{CERUSE,reimers2019sentence} and even authors, with contributions like Usr2Vec \cite{amir2017quantifying}, Aut2Vec \cite{ganguly2016author2vec},  and DGEA \cite{gourru2022dgea}. The main drawback of these models is that they were shown by \cite{terreau2021writing} to mainly focus on topics rather than on stylistic features of the text. It turns out that capturing writing style can be of much interest for some applications.

When working with literacy data or for forensic investigation \cite{yang_aa_forensic}, practitioners are generally interested in detecting similarities in writing style regardless of the topics covered by the authors.
The author style can be defined as every writing choice made without semantic information, often study through various linguistic and syntactic features. As demonstrated by \cite{terreau2021writing}, most author embedding techniques rely on the semantic content of documents: a poem and a fiction writing on flowers will be placed closer in the latent space, regardless of their strong differences in sentence construction, structure, etc.

\begin{figure}[t]
  \centering
  \includegraphics[width=\columnwidth]{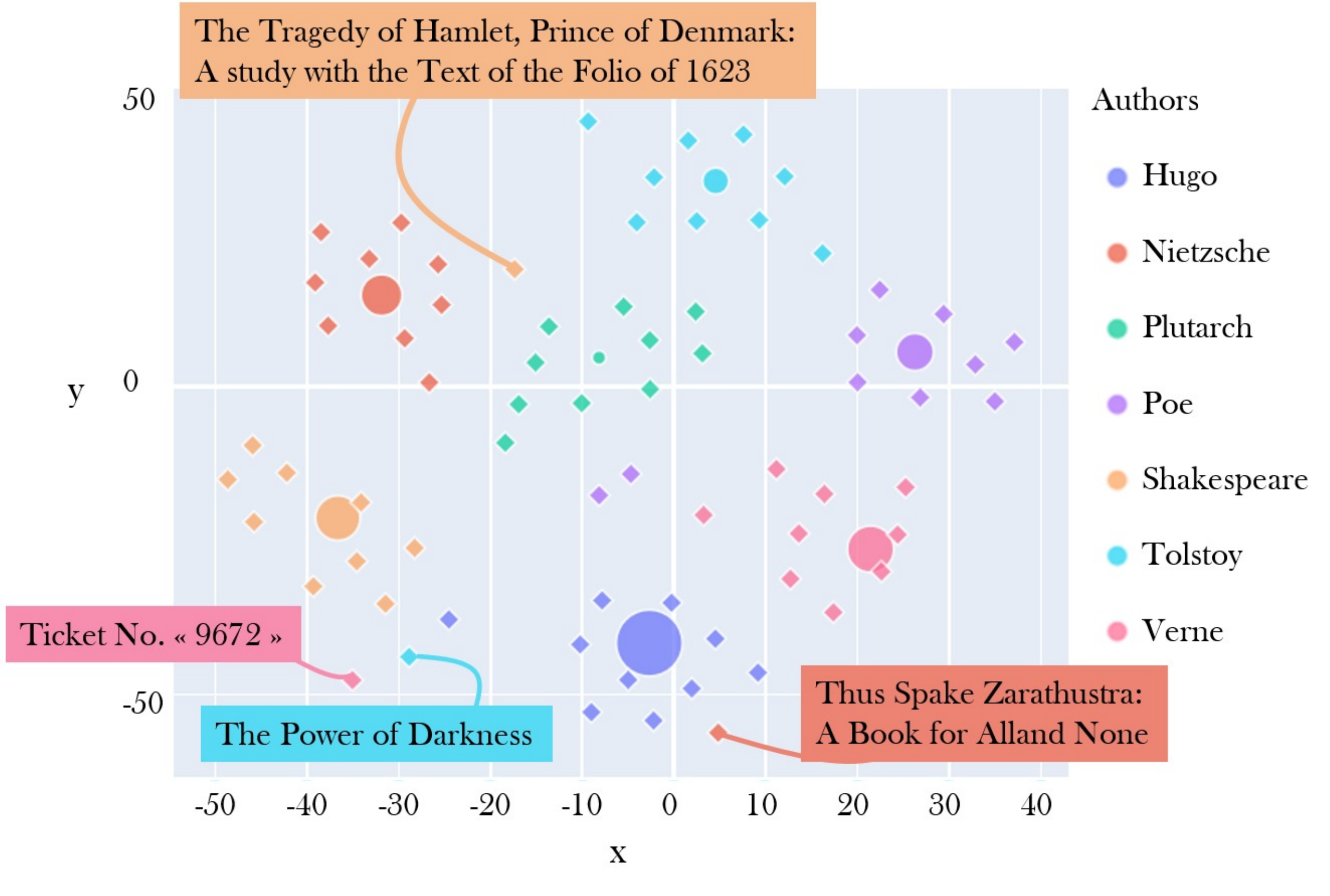}
  \caption{\textbf{Author and book representations from R-PGD.}\hspace{\textwidth} We here present a 2D projection with T-SNE of \modelname\ documents and authors embeddings on R-PGD. Books are represented with diamond, authors with dot. The bigger the dot, the bigger the author variance learnt.}\label{fig:embeddings}
\end{figure}

As an answer to these limitations, we propose a new model that builds a representation space which captures writing style by using stylistic metrics as additional input features. We follow \cite{VIBsentence} and leverage the Variational Information Bottleneck (VIB) framework \cite{alemi2016deep}, that was shown to outperform the classical pointwise contrastive training. More precisely, we propose to use it to fine tune a pretrained document encoder (such as \cite{ CERUSE}) and author representations on an authorship attribution task. This is, to our knowledge, the first time that this framework is applied to author representation learning. Then, we add an additional term in the objective function to enforce the representations to capture stylistic features. We name this new model \modelname. Using pretrained models allows to benefit from accurate intermediate text representations, built on ready-to-use language resources. In Figure \ref{fig:embeddings}, we present a subset of authors from the Project Gutenberg and the representation of the documents they wrote. The size of author's vector is proportional to its variance, learnt by using the VIB framework. As expected, some outlier productions from authors in term of style (e.g., Thus Spake Zarathustra from Nietzsche) lie closer in the representation space to books of the same genre. More precisely, our model allows 1)~to capture author and document style, 2)~to build an interpretable representation space to be used by researchers in linguistic, literature and public at large, 3)~to predict stylistic features such as readibility index, NER frequencies, more accurately than every existing neural based methods, 4)~accurately identify document's author, even when they are unknown. 

After a presentation of related works, we introduce the theoretical foundations of the VIB framework, we then describe our model and how it is optimized. In the last section, we present experimental results on two tasks: author identification and stylistic features prediction. Our experiments demonstrate that our model outperforms or matches existing author embedding methods, in addition to being able to infer representations for unseen documents, measure semantic uncertainty of authors and documents, and capture author stylistic information.

\section{Related Works}

\subsection{Author Embedding Models}

Word embedding, popularized by \cite{mikolov2013distributed}, was then extended to document embedding by the same authors. More recent works \cite{CERUSE} propose different aggregation functions of word embeddings, based on LSTM, Transformers, and Deep Averaging Networks, to build (short) document level representations. The aggregations is learnt through classification or document pairing. More recently, \cite{reimers2019sentence} proceed in a similar way by fine tuning a BERT model \cite{devlin2019bert}.
 
There are also specific works focusing on author embeddings. The Author Topic Model (ATM) \cite{rosen2004author} is a hierarchical graphical model, optimized through Gibbs sampling. It produces a distribution over jointly learnt topic factors that can be used as author features. Aut2vec \cite{ganguly2016author2vec} allows to learn representations of authors and documents that can separate true observed pairs and negative sampled (document, author) pairs. The distance between two representations modifies an activation function producing a probability that the pair is observed in the corpus. This approach concatenates two sub models: the Link Info model, which takes pairs of collaborating authors, and the Content Info model, which uses pairs of author and documents. It cannot infer representations of unseen documents and authors: the embeddings are parameters of an embedding layer. The Usr2vec model \cite{amir2017quantifying} learns author representation from pretrained word vectors. Authors use the same objective than \cite{mikolov2013distributed}, and add an author id to learn the representations. 

\subsection{Writing Style-oriented Embedding Models}

Although there is no consensus definition of writing style, it has always been a widely addressed research topic. In computational linguistic, the approach of \cite{karlgren2004} is often cited as a reference and gives the following definition: \textit{``Style is, on a surface level, very obviously detectable as the choice between items
in a vocabulary, between types of syntactical constructions, between the various ways a text can be woven from the material it is made of.''}, and the author to conclude further to the \textit{``impossibility of drawing a clean line between meaning and style''}. That's why style is commonly defined as every writing choice without semantic information. 

Based on this definition, it is hard, if not impossible, to produce a clear annotated dataset that classifies different writing styles. The workaround in most studies is to identify the most useful stylistic features to associate an author to its production.
It starts in the 19th century with \cite{Mendenhall237} and the most basic features (e.g., word and punctuation frequencies, hapax legomena, average sentence length). More recent work focuses on function words frequencies \cite{Zhao2005}, hybrid variables such as character n grams \cite{Stamatatos2013, schwartz-etal-2013-authorship}, or even part-of-speech (POS) and name entity recognition (NER) tag frequencies, using authorship prediction as evaluation.

Several methods try to use these stylistic features to learn document representations. For example, \cite{Maharjan2019} use Doc2Vec on documents of character trigrams annotated regarding their position in the word or if they contain punctuation (NGRAM Doc2Vec). According to the authors, it allows to capture both content and writing style. In an other work, words and POS tags embeddings are learnt together before passing them through a CNN to get a sentence representation \cite{styleneuraldoc}. Then these sentences are fed into an LSTM with a final attention layer to compute document representation. This model is trained on the authorship attribution task.

Some works claim to capture this information in an unsupervised manner. DBert-ft \cite{hay-etal-2020-representation} fine-tunes DistilBERT on the authorship attribution task, assuming that an author writing style must be consistent over its documents, and thus, that this task allows to build a ``stylometric latent space'' when the model is trained on a reference set. Yet, for all above models, no author representation is explicitely learnt.

\section{Our model: \modelname}

\begin{figure}[t]
  \centering
  \includegraphics[width=\columnwidth]{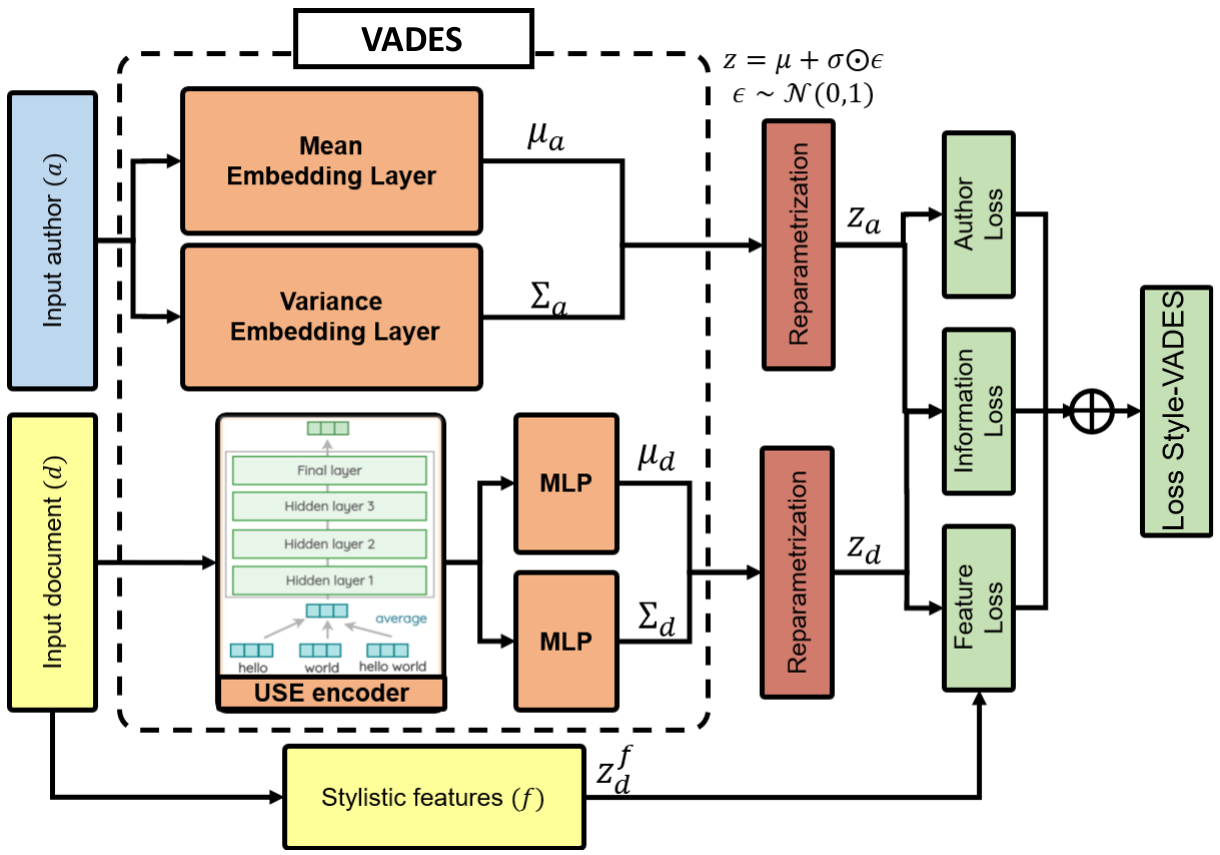}
  \caption{\textbf{\modelname\ in one picture.} \hspace{\textwidth} We draw a single representation $z_d$ using the reparametrization trick. Authors mean and variance are trainable parameters (embedding layers). $L_{\modelname}$ computes the probability of the author/document pair to be observed, plus a regularization term and a stylometric features-based loss, see Eq.\ref{eq:lv}.}\label{fig:vador}
\end{figure}

\subsection{Goal and VIB Framework}

We deal with a set of documents, such as literature or blog posts. We assume each document is written by one author. 
Each document of indice $d$ is preprocessed to extract a vector $z_{d}^{f}$ of $r = 300$ stylistic features following \cite{terreau2021writing}.
 
Our goal is threefold: i)~We want to build author and document representations \emph{in the same space} $\mathbb{R}^r$ such that their proximity captures their stylistic similarity (Figure \ref{fig:embeddings}), ii)~We want to learn a measure of variability in style for each document and author , and iii)~We want our model to incorporate an on-the-shelf pre-trained text encoder such as Sentence-BERT or USE to benefit from their complex language understanding, fine-tuned on the dataset at hand using the objective we have just defined. To do that, we build an architecture based on the Variational Information Bottleneck (VIB) framework.

The VIB framework is a variational extension of the Information Bottleneck principle \cite{tishbyinformation} proposed by \cite{alemi2016deep}. The general objective function is, for a set of observations $x$, to associate labels $y$ and latent representations $z$ of these observations:
\begin{equation}
\label{eq:vib}
    \arg\max\limits_z I(z, y)-\beta I(z, x), 
\end{equation}
where I is the well-known Mutual Information measure, defined as:
\begin{equation}
    I(x,y) = \int \int p(x,y) \log \frac{p(x,y)}{p(x)p(y)} d_x d_y.
\end{equation}
Information Bottleneck aims at maximally compressing the information in $z$, such that $z$ is highly informative regarding the labels, i.e. $z$ can be used to predict the labels $y$. With $y$ being a set of relevant stylistic features, we would like to maximize the stylistic information captured by the representation, while minimizing the semantic one. $ \beta \geq 0 $ is a hyper-parameter that controls the balance between the two sub-objectives. 

In this approach, $p(z|x)$ (the ``encoding law'') is defined by modeling choices. Most of the time, the mutual information is intractable. We then obtain a lower bound of Eq.\ref{eq:vib} by using variational approximations thanks to \cite{alemi2016deep}:
\begin{equation}
\label{borne}
    -L_{vib} = \mathbb{E}[\log q(y|z)] - \beta KL(p(z|x)||q(z))
\end{equation}
\noindent where $q(y|z)$ is a variational approximation of $p(y|z)$ and $q(z)$ approximates $p(z)$. Maximizing Eq.\ref{borne} leads to increasing Eq.\ref{eq:vib}. 

\subsection{VIB for Embedding with Stylistic Constraints}

\cite{oh2018modeling} propose to use this framework to learn probabilistic representations of images. They leverage an instance of this framework based on siamese networks with a (soft) contrastive loss objective function, to separate positive observed pairs of images ($y = 1$) and negative examples ($y = 0$). We extend this model to document and author embedding with stylistic constraint. Each author $a$ (resp. document $d$) is associated to a stochastic representation $z_{a}$ (resp. $z_{d}$) that is unobserved (i.e., latent). Additionally, each document is associated to a stylistic feature vector $z_d^f$ that is beforehand extracted from the corpus with usual NLP toolkits. We assume that the dimensions of $z_a$, $z_d$ and $z_d^f$ are the same ($r$).

We build a set of pairs $(a,d)$ with label $y_a=1$ if $a$ wrote $d$. We additionally draw $k$ negative pairs $(a',d)$ for each observed pair, associated with label $y_a=0$, where $a'$ is \emph{not} an author of $d$. The encoding laws ($p(z|x)$) for authors and documents are normal laws. To capture stylistic information, we also build a set of pairs $(d,d)$ with label $y_f=1$ and we draw $k$ negative pairs $(d, d')$ for each observed pair, associated with label $y_f=0$. These pairs are used to train the stylistic objective : the representation $z_d$ of a document should be close to its feature vector $z_{d}^{f}$.

We learn the following parameters for each author $a$: mean $\mu_a$ and diagonal variance matrix with diagonal $\sigma_a^2$ (these are embedding layers). For a document $d$, we use a trainable text encoder to map a document's content to a vector $d_0 \in \mathbb{R}^{r_0}$. We then build the document mean $\mu_d = f(d_0) \in \mathbb{R}^{r}$ and diagonal variance matrix with diagonal $\sigma_d^2 = g(d_0) \in \mathbb{R}^{r}$. As we will show later, the dimension $r$ should match the number of stylistic features to gain in comprehension of the learning space, but the text encoder can output vectors of any dimension (here, $r_0$). Following \cite{alemi2016deep,oh2018modeling}, $f$ and $g$ are neural networks. We give more details on $f$, $g$ (the ``encoding functions''), and the text encoder later.

Following \cite{oh2018modeling}, the probability of a label is the soft contrastive loss:
\begin{equation}
\begin{split}
    q(y_a = 1 | z_{a}, z_{d}) &=\sigma(-c_a|| z_{a}-z_{d}||_{2}+e_a)\\
q(y_f = 1 | z_{d}, z_{d}^{f}) &=\sigma(-c_f|| z_{d}-z_{d}^{f}||_{2}+e_f),
\end{split}
\end{equation}

\noindent where $\sigma$ is the sigmoid function, $ c_a, c_f> 0 $ and $e_a,e_f \in \mathbb{R}$. We introduce an additional parameter $\alpha \in [0,1]$ to control the importance given to the features and to the authorship prediction objective. We can define the loss function (to minimize) based on the VIB framework as follows:
\begin{equation}\label{eq:lv}
\begin{split}
    \mathcal{L}=&  -(1-\alpha)\mathbb{E}_{p(z_{a}|x_{a}),p(z_{d}|x_{d})}[\log q(y_a | z_{a}, z_{d})]\\ &-\alpha\mathbb{E}_{p(z_{d}|x_{d})}[\log q(y_f | z_{d},z_{d}^{f})] \\
    &+ \beta \left( KL(p(z_{a}|x_{a})||q(z_{a})) + KL(p(z_{d}|x_{d})||q(z_{d})) \right)
\end{split}
\end{equation}

Here, $\alpha = 0$ will produce representations that well predict the author-document relation but will not capture the stylistic features of the documents, as shown by \cite{terreau2021writing}. With $\alpha = 1$, on the contrary, the model will simply bring document embeddings closer to their feature vectors. Hence, the value of $\alpha$ needs to be carefully tuned \emph{on the dataset}, regarding if the corpus is writing style specific or not thanks to domain knowledge.

Eventually, computing the expected values in Eq.(\ref{eq:lv}) is intractable for a wide range of encoders. We therefore approximate it by sampling $L$ examples by observation (here, a triplet document, author, feature vector), following $p(z|x)$ as done in \cite{oh2018modeling}. We get (the same goes for feature vector/documents pairs) : 
\begin{equation}\label{eq:esperance}
\begin{split}
    \mathbb{E}[\log q\left(y_a| z_{a}, z_{d}\right)] &\approx \frac{1}{L}\sum_{l=1}^{L}{\log q(y_a| z^{(l)}_{a}, z^{(l)}_{d})}\\
\end{split}
\end{equation}
We then use the reparametrization trick, following what is done in VAE \cite{kingma2013auto}:
\begin{equation*}\label{eq:tirages}
    z^{(l)}_{a} = \mu_a + \sigma_a \odot \epsilon\text{,} \hspace{0.5em} z^{(l)}_{d} = \mu_d + \sigma_d \odot \epsilon \hspace{0.5em} \text{with} \hspace{0.5em} \epsilon \sim \mathcal{N}(0,1)
\end{equation*}
This loss can now be minimized using backpropagation. In Figure \ref{fig:vador}, we show a schematic representation of our model, called \textbf{\modelname} for Variational Author and Document Representations with Style.

\subsection{Encoding Functions and Choice of the Encoder}
\label{subsec_encoder}

The entering bloc of our model for documents is a text encoder, mapping a document in natural language to a vector in $\mathbb{R}^r$. Many deep architectures could be used here and trained from scratch. Nevertheless, we propose to use a pretrained text encoder. 

Models that are pretrained on large datasets are now easily available online\footnote{e.g., \url{https://huggingface.co/models}}. They have been proved successful on many NLP tasks with a simple fine-tuning phase (the only constraint being to avoid catastrophic forgetting). Additionally, the VIB framework allows to naturally introduce a pretrained text encoder as shown by \cite{VIBsentence}. The encoder's output should then be mapped to document mean and variance. Both \cite{VIBsentence,gourru2022dgea} map the text encoder output to the document's mean (the $f$ function) and variance (the $g$ function) using a Multi Layer Perceptron (MLP). This approach is simple, and fast. In our experiments, we build $f$ and $g$ as two-layer MLP with \emph{tanh} and linear activation with same input and intermediate dimensions ($r_0$). Note that the output dimension of $f$ and $g$ should be the same as the number of stylistic features ($r$).

 Several constraints arise regarding the pretrained encoder itself. We would like our model to be able to capture stylistic information from a given document. As shown in \cite{bertlookat, terreau2021writing}, state-of-the-art models trained on large datasets already capture complex grammatical and syntactic notions in their representations, and therefore have the explanatory power requested for our objective. Moreover, our model must be able to deal with long text as it will be used in a literary context. Processing novels, dramas, essays, where writing style interferes the most. This is a serious problem: for example, the widely used BERT model is limited to 512 tokens. Alternative models such as \cite{bigbird} allow to apply transformers to long documents. To circumvent this issue, we use the Deep Averaging Network implementation of the Universal Sentence Encoder (USE) from \cite{CERUSE}. It has several advantages over the latter works: it gives no length constraint, it is faster than transformer-based methods and it outperforms Sentence-BERT on stylistic features prediction \cite{terreau2021writing}. The test of other encoder models is left to future works. Finally, note that our model is language agnostic (as it depends on a out-of-the-box text encoder) and can infer representations for unseen documents.

\section{Authorship Attribution Datasets}
\subsection{IMDb Corpus}
The IMDb (Internet Movie Database) corpus is one of the most used ones regarding the authorship attribution task. It was introduced by \cite{imdbcorpus} and is composed of $271,000$ movie reviews from $22,116$ online users. However, most of the works are evaluated on the reduction of this dataset to only 62 authors with $1000$ texts for each (IMDb62). Thus, we benchmark our model on IMDb62. As shown later, the task of authorship attribution on this corpus is more or less solved, due to the low number of authors.

\subsection{Project Gutenberg Dataset}
The Project Gutenberg is a multilingual library of more than 60,000 e-books for which U.S. copyright has expired. It is freely available and started in 1971. We gathered the corpus using \cite{spgd}. Most of the books are classical novels, dramas, essays, etc. from different eras, which is relevant when studying writing style and represents quite well our context of application. To keep the most authors possible, we randomly sample 10 texts for each author with such a production, leaving 664 authors in our Reduce Project Gutenberg Dataset (R-PGD) (10 times more than IMDB). To be able to deal with such works, we only keep the 200 first sentences of each book.

\subsection{Blog Authorship Corpus}
This dataset is composed of 681,288 posts from 19,320 authors gathered in the early 2000s by \cite{bac_corpus}. There are approximately 35 posts and 7,250 words by user. We only take 500 bloggers with at least 50 blogposts to build our reduced dataset of the Blog Authorship Corpus (R-BAC). This dataset is also used in several authorship attribution benchmark, only keeping the top 10 or 50 authors with most productions. We will also test our model on these extraction of the corpus.

\begin{figure*}[t]
  \centering
  \includegraphics[width=\textwidth]{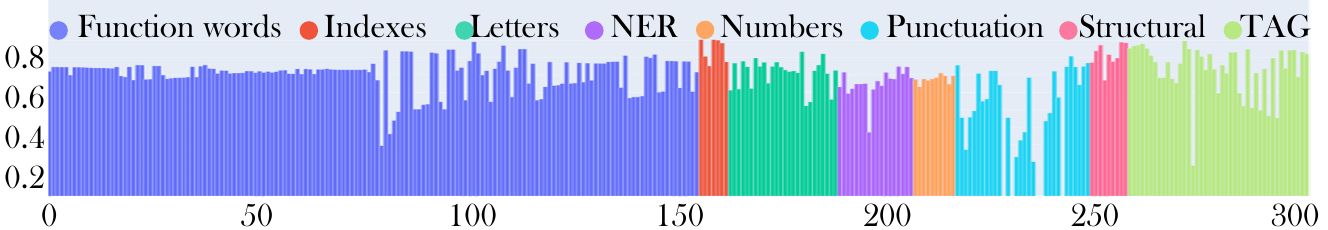}
  \caption{\textbf{Correlation score between $i^{th}$ embedding coordinates and $i^{th}$ stylistic feature for \modelname\ representation on R-PGD.} \hspace{\textwidth} A few values in the Punctuation categories are null as they were not found anywhere in the corpus.}\label{fig:correlation}
\end{figure*}

These two last datasets (PGD and BAC) represent two common uses of author embedding (classic literature and web analysis) with a large number of authors. Usual datasets for authorship attribution (CCAT50, NYT, IMDb62) contain far less classes, further from our context of a web extracted corpus (from Blogger or Wordpress for example)... They are also stylistically and structurally different, allowing to evaluate our approach on various textual formats. For each dataset, we perform a 80/20 train-test stratified split.

\begin{figure*}
  \centering
  \includegraphics[width=\textwidth]{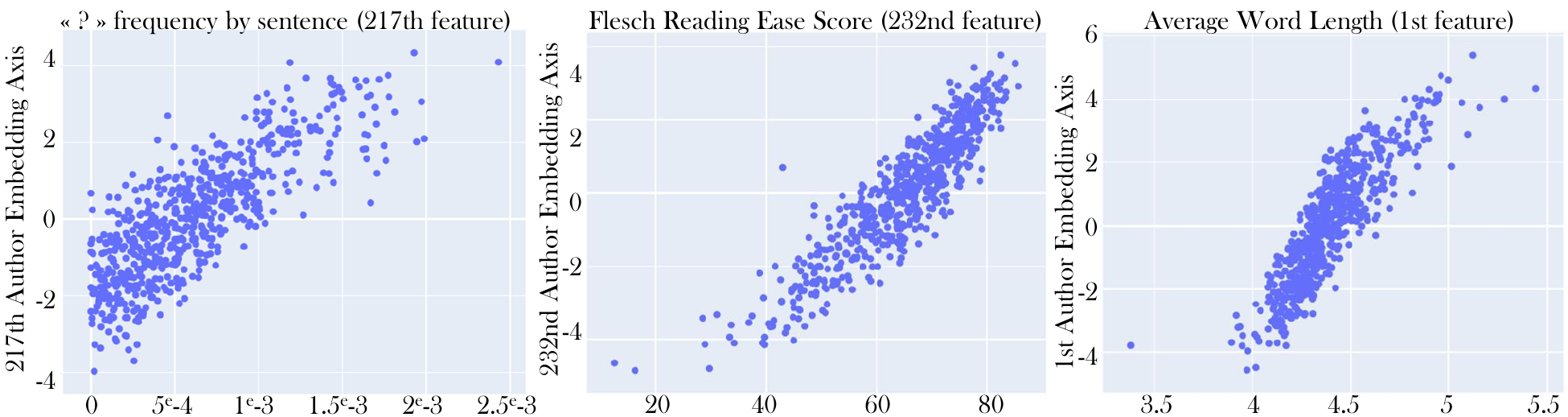}
  \caption{\textbf{$i^{th}$ embedding axis against $i^{th}$ stylistic feature for each author representation, for a selection of 4 given features} \hspace{\textwidth}
  We can see correlation between each feature and their respective embedding axis.}\label{fig:correlation_selection}
\end{figure*}

\begin{table}
\centering
\begin{tabular}{llll}
\toprule
\multicolumn{4}{c}{\textbf{Datasets statistics}}\\
\textbf{Dataset} & \textbf{Authors} & \textbf{Avg. Tokens} & \textbf{Avg. Texts}\\
\toprule
IMDb62 & 62 & $341 (\pm 223)$ & $1000 (\pm 0)$ \\
\midrule
BAC10 & 10 & $91 (\pm 184)$ & $2350 (\pm 639)$ \\
BAC50 & 50 & $98 (\pm 167)$ & $1466 (\pm 562)$ \\
R-BAC & 500 & $243 (\pm 342)$ & $50 (\pm 0)$ \\
\midrule
R-PGD & 664 & $2315 (\pm 961)$ & $10 (\pm 0)$ \\
\bottomrule
\end{tabular}
\label{dataset_statistics}
\caption{\textbf{Descriptive statistics for the 3 datasets and their decomposition.} \hspace{\textwidth} BAC : Blog Authorship Corpus, PGD : Project Gutenberg Dataset.}
\end{table}

\begin{table}
\centering
\begin{tabular}{ll}
\toprule
\multicolumn{2}{c}{\textbf{Hyperparameter grid search}}\\
\textbf{Hyperparameter} & \textbf{Grid}\\
\toprule
\# negative pairs & \{1, 5, \textbf{10}, 20\}\\
\midrule
Monte Carlo sampling & \{1, 5, \textbf{10}, 20\}\\
\midrule
Learning rate & \{1e-2, \textbf{1e-3}, 1e-4, 1e-5\} \\
\midrule
$\beta$ & \{1e-1, 1e-2, ..., \textbf{1e-12}\} \\ 
\midrule
Feature loss & \{L2, \textbf{Cross-Entropy}\} \\
\bottomrule
\end{tabular}
\caption{\label{gridsearch}
\textbf{Grid search used for hyperparameter selection.} \hspace{\textwidth} Selected value in bold.
}
\end{table}

\section{Experiments}

\subsection{Parameter Setting and Competitors}

In this section, we present implementation details for our method and competitors. For the encoder functions $f$ and $g$, we use the architectures presented in the previous section with batch normalization and dropout equal to 0.2 with L2 regularization ($1e-5$). Grid-search parameters are detailed in Table \ref{gridsearch}. For $L$, we obtain a good trade-off between accuracy and speed with $L=10$, as we quickly reach a plateau of performance when increasing its value. We can summarize the tuning of $\alpha$ as follows:
\begin{itemize}
    \item $\alpha=0$ implies no feature loss and stylistic information,
    \item $\alpha =0.5$ gives the same importance to feature loss and author loss,
    \item $\alpha = 0.9$ pushes feature loss to boost style detection.
\end{itemize} 
We train the model for 15 epochs on R-PGD and R-BAC, and for 5 epochs on IMDB, BAC10 and BAC50 as the number of authors is around ten times smaller. We use a partition of 2 GPUs V100. On a single GPU, training the model on the R-PGD dataset takes around 10 hours. 
In the following section, we report the results for the best version of \modelname\ only. As an ablation study, to justify the use of both the VIB framework and stylistic features, we compare our model with and without these components (respectively called \textit{\modelname\ no-VIB} and 
 \textit{\modelname\ ($\alpha = 0$))}. The code is available on github and will be shared if the paper is accepted. All the datasets are available online.

We compare our model with several baselines. We use \cite{Maharjan2019} (NGRAM Doc2Vec), a simple average based version of USE \cite{CERUSE} (a document representation is built from the average of its sentence encoding, and an author representation is an average of its documents). We also compare our approach to DBert-ft \cite{hay-etal-2020-representation}, a document embedding method where DistilBERT is fine-tuned on the authorship attribution task. The author embeddings are built by averaging the representations of the documents it wrote. We use the parameters detailed in the authors' implementation\footnote{\url{https://github.com/hayj/DBert-ft}}.

\subsection{Evaluation Tasks}

We first evaluate the baselines and \modelname\ regarding how well each method captures writing style. As writing style is a complex and a still discussed notion, there is no supervised dataset to evaluate how a model can grasp it. We therefore use a proxy task that consists in predicting stylistic feature from the latent representations. We follow the experimental protocol of \cite{terreau2021writing}. The stylistic features are extracted using spacy word and sentence tokenizer, POS-tagger and Name Entity Recognition, spacy English stopwords and nltk CMU Dictionary. For each author, we aim to predict the value of all stylistic features from their embeddings. Each feature is standardized before regression. We use an SVR with Radial Basis Function (rbf) kernel as it offers both quick training time and best results among other kernels in our experiments. We evaluate models using Mean Squared Error (MSE) following a 10-fold cross validation scheme.

Secondly, we perform authorship attribution, the task of predicting the author of a given document. We compare \modelname\ with several other authorship attribution methods even though they do not necessarily perform representation learning. Each dataset is split into train and test sets with a 80/20 ratio. For our model, we repeated 5 times the evaluation scheme. For embedding method without classification head, we associate each document with its most plausible author using cosine similarity. We use accuracy to evaluate these results (the percentage of correctly predicted authors out of all data points).

\begin{table*}[t]
\centering
\resizebox{\textwidth}{!}{
\begin{tabular}{lllllllll}
\toprule
 & \multicolumn{8}{c}{\textbf{Average MSE Regression Score along with standard deviation (SVR Model) on R-PGD dataset}}\\
\textbf{Embedding} & \textbf{Letters} & \textbf{Numbers} & \textbf{Structural} & \textbf{Punctuation} & \textbf{Func. words} & \textbf{TAG} & \textbf{NER} & \textbf{Indexes}\\
\toprule
\textbf{Content-Info} & 0.67 (0.17) & 0.88 (0.12) & 0.55 (0.19) & 0.68 (0.16) & 0.72 (0.19) & 0.65 (0.17) & 0.74 (0.14) & 0.50 (0.16)\\
\midrule
\textbf{Ngram Doc2Vec} & 0.63 (0.20) & 0.88 (0.12) & 0.51 (0.20) & 0.58 (0.21) & 0.68 (0.19) & 0.59 (0.19) & 0.71 (0.14) & 0.45 (0.15)\\
\midrule
\textbf{USE} & 0.61 (0.27) & 0.86 (0.09) & 0.34 (0.18) & \underline{0.59 (0.26)} & 0.65 (0.24) & 0.45 (0.29) &  0.65 (0.17) & 0.27 (0.15)\\
\midrule
\textbf{DBert-ft} & 0.79 (0.16) & 0.92 (0.09) & 0.65 (0.15) & 0.82 (0.17) & 0.84 (0.13) & 0.74 (0.14) & 0.84 (0.08) & 0.60 (0.14)\\
\midrule
\textbf{\modelname\ no-VIB (0.5)} & 0.55 (0.23) & 0.67 (0.11) & 0.32 (0.14) & 0.66 (0.27) & 0.58 (0.21) & 0.44 (0.27) & 0.62 (0.16) & 0.24 (0.14)\\
\midrule
\textbf{\modelname\ (0.0)} & 0.84 (0.24) & 0.91 (0.12) & 0.66 (0.13) & 0.85 (0.18) & 0.91 (0.15) & 0.71 (0.23) & 0.88 (0.09) & 0.61 (0.16)\\
\midrule
\textbf{\modelname\ (0.5)} & \underline{0.50 (0.22)} & \underline{0.60 (0.11)} & \underline{0.28 (0.14)} & 0.62 (0.27) & \underline{0.53 (0.21)} & \underline{0.40 (0.27)} & \underline{0.58 (0.15)} & \underline{0.20 (0.11)}\\
\midrule
\textbf{\modelname\ (0.9)} & \textbf{0.47 (0.22)} & \textbf{0.53 (0.10)} & \textbf{0.26 (0.13)} & \textbf{0.59 (0.28)} & \textbf{0.50 (0.21)} & \textbf{0.39 (0.26)} & \textbf{0.56 (0.15)} & \textbf{0.19 (0.10)}\\
\bottomrule
\toprule
 & \multicolumn{8}{c}{\textbf{Average MSE Regression Score along with standard deviation (SVR Model) on R-BAC dataset}}\\
\textbf{Embedding} & \textbf{Letters} & \textbf{Numbers} & \textbf{Structural} & \textbf{Punctuation} & \textbf{Func. words} & \textbf{TAG} & \textbf{NER} & \textbf{Indexes}\\
\toprule
\textbf{Content-Info} & 0.80 (0.15) & 0.85 (0.07) & 0.62 (0.23) & 0.92 (0.09) & 0.87 (0.12) & 0.90 (0.05) & 0.93 (0.07) & 0.70 (0.29)\\
\midrule
\textbf{Ngram Doc2Vec} & 0.77 (0.16) & 0.88 (0.05) & 0.67 (0.16) & \underline{0.78 (0.13)} & 0.84 (0.12) & 0.82 (0.09) & 0.86 (0.11) & 0.67 (0.13)\\
\midrule
\textbf{USE} & \underline{0.67 (0.25)} & \underline{0.83 (0.05)} & \underline{0.45 (0.20)} & 0.78 (0.17) & \underline{0.81 (0.17)} & \underline{0.63 (0.21)} & \underline{0.80 (0.17)} & \underline{0.38 (0.18)}\\
\midrule
\textbf{DBert-ft} & 1.05 (0.09) & 1.05 (0.07) & 1.01 (0.05) & 0.98 (0.22) & 1.05 (0.09) & 0.95 (0.19) & 0.91 (0.20) & 1.03 (0.07)\\
\midrule
\textbf{\modelname\ (0.9)} & \textbf{0.52 (0.23)} & \textbf{0.55 (0.09)} & \textbf{0.31 (0.17}) & \textbf{0.76 (0.22)} & \textbf{0.67 (0.20)} & \textbf{0.57 (0.20)} & \textbf{0.73 (0.18)} & \textbf{0.32 (0.20)}\\
\midrule
\bottomrule
\end{tabular}}
\caption{\textbf{Feature prediction on R-PGD and R-BAC.} \hspace{\textwidth} MSE score (standard deviation in parenthesis) on the prediction of stylistic features from author embedding on the R-BAC dataset using SVR. The 300 stylistic features are grouped by families. In bold the best scores for each axis. Our model ($\alpha$ value in parenthesis) performs best with $\alpha = 0.9$.
}\label{blog-results}
\end{table*}

\begin{table}
\centering
\begin{tabular}{llll}
\toprule
& \textbf{IMDb62} & \multicolumn{2}{c}{\textbf{Blog Authorship Corpus}}\\
\textbf{Approach} & \textbf{62 authors} & \textbf{10 authors} & \textbf{50 authors} \\
\midrule
Stylistic features + LR & 88.2 (0.1) & 40.9 (0.2) & 28.4 (0.2) \\
LDA+Hellinger* \cite{LDAHellinger} & 82 & 52.5 & 18.3\\
Impostors* \cite{impostors} & x & 35.4 & 22.6 \\
Word Level TF-IDF* & 91.4 & x & x \\
CNN-Char* \cite{cnn-char} & 91.7 & 61.2 & 49.4\\
C.Att + Sep.Rec.* \cite{compe_att_aa} & 91.8 & x & x \\
Token-SVM* \cite{imdbcorpus} & 92.5 & x & x \\
SCAP* \cite{frantzeskou_aa} & 94.8 & 48.6 & 41.6\\
Cont. N-gram* \cite{sari-etal-2017-continuous} & 94.8 & 61.3 & 52.8\\
(C+W+POS)/LM* \cite{kampetal_aa} & 95.9 & x & x\\
N-gram + Style* \cite{Sari2018} & 95.9 & x & x\\
N-gram CNN* \cite{zhang-etal-2018-syntax} & x & 63.7 & 53.1 \\
Syntax CNN* \cite{zhang-etal-2018-syntax} & \underline{96.2} & 64.1 & 56.7\\
DBert-ft \cite{hay-etal-2020-representation} & \textbf{96.7 (0.2)} & \underline{64.3} (0.2) & \underline{58.5} (0.2)\\
BertAA* \cite{fabien-etal-2020-bertaa}& 93.0 & \textbf{65.4} & \textbf{59.7}\\
\midrule 
\modelname\ no-VIB (0.5) & 91.3 (0.1) & 60.9 (0.2) & 50.2 (0.2)\\
\modelname\ (0.0) & 94.9 (0.2) & 62.6 (0.2) & 52.4 (0.2) \\
\modelname\ (0.1) & 95.6 (0.2) & 63.8 (0.2) & 53.8 (0.2) \\
\bottomrule
\end{tabular}
\caption{\textbf{Authorship Attribution accuracy on IMDb62 and Blog Authorship Corpus} \hspace{\textwidth}  Results with * are gathered from other papers, x is for missing results on a given dataset. Best model in bold and second underlined.
We here compare our model (in parenthesis $\alpha$ value) with several authorship attribution models. Our model compete with SOTA model while learning meaningful representations regarding writing style for documents and authors.
}\label{imdb62-results}
\end{table}

\subsection{Results on capturing writing style}

As explained earlier, we use the author embeddings to perform regression and predict each stylistic features. As shown in Table \ref{imdb62-results}, only using a simple logistic regression on these stylistic features allows to reach decent scores in authorship attribution, close to these of Universal Sentence Encoder, which is a state-of-the-art method in sentence embedding. As they contain strictly no topic information, it demonstrates how good they are as a proxy of writing style. Thus, a model able to capture them is able to capture writing style.

Results on the style MSE metric are shown in Table \ref{blog-results}. As expected, our model easily outperforms every baseline on all axes. DBert-ft, only trained on the authorship attribution objective performs the worst. Even though this approach is based on fine-tuned language models which already capture syntactic and grammatical notions \cite{bertlookat}, this is not the information that seems to be retained by the network when trained on the author attribution task. This is consistent with what was shown in \cite{terreau2021writing}. The models may mainly focus on the semantic information to predict author-document relation. Interestingly, we observe that a simple average of USE representations performs quite well, which confirms that it can successfully capture complex linguistic concepts. \modelname\ is guided by the feature loss to do so.

\begin{figure}
\centering
\includegraphics[width=0.95\columnwidth]{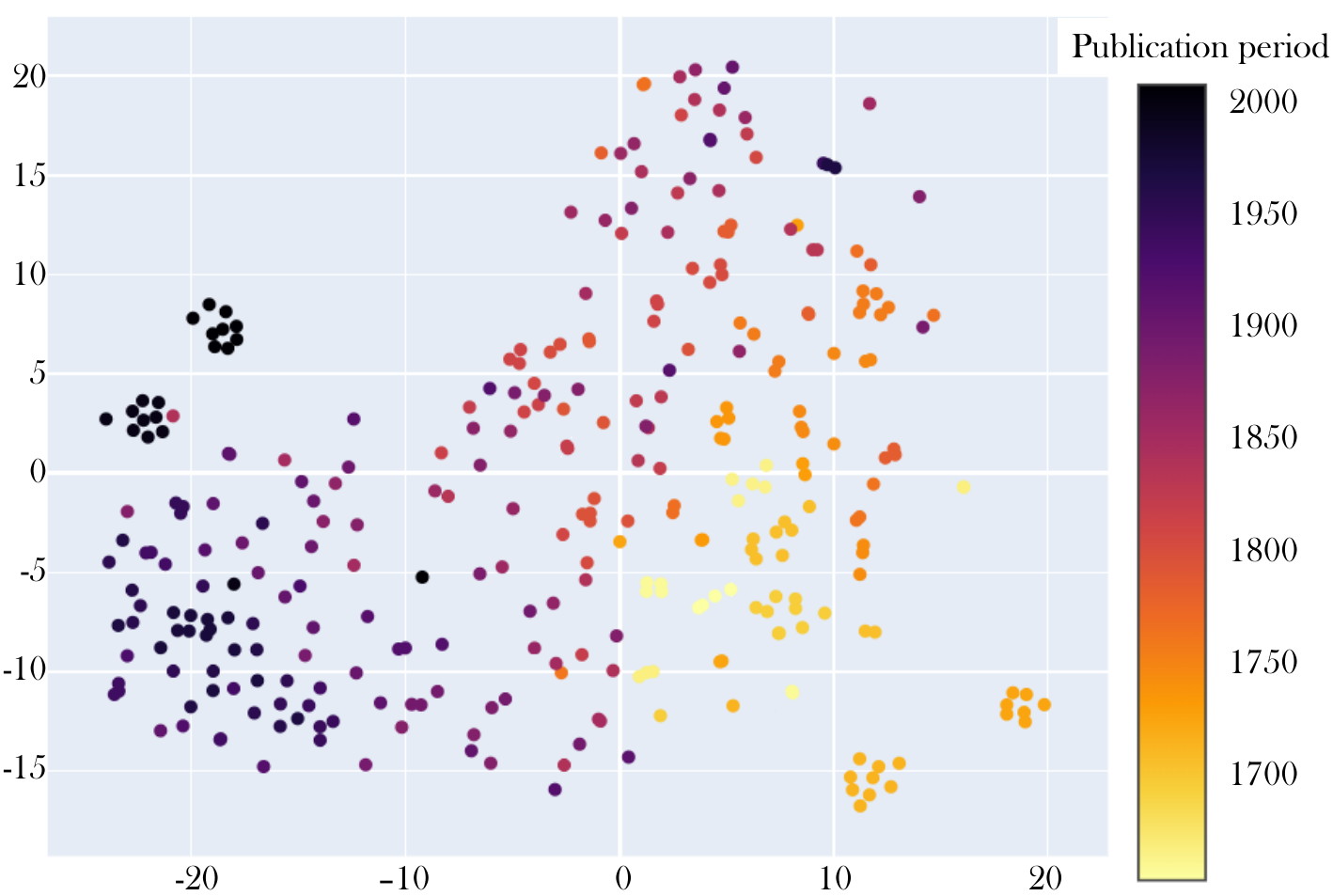}
\caption{\label{time_embeddings}
\textbf{Books representations from Project Gutenberg and their writing period.} \hspace{\textwidth} We sampled 10 R-PGB books by decades starting in 1650 and present here a 2D T-SNE projection of their VADES embeddings.}
\end{figure}

On a qualitative note, we present two additional visualisations to underline the strong advantage of \modelname\ for linguistic and stylistic applications. In Figure~\ref{time_embeddings}, we present a T-SNE 2D projection of the books of the R-PGD dataset colored by their publication year. A clear color gradient appears, demonstrating that our model can grasp the evolution of writing style through the last centuries. Figure \ref{fig:embeddings} shows a toy example of a T-SNE 2D projection of well-known authors from the R-PGD dataset and their books (we use $\alpha = 0.5$). The objects are distributed in the space across clear author specific clusters. The most interesting observation is related to documents that are outside of their author cluster: \textit{Thus Spake Zarathustra: A Book for All and None} by Nietzsche is a philosophical poem, closer to Hugo, while the rest of its production is mostly essays. The same conclusion goes with \textit{The Power of Darkness} by Tolstoï, a 5 acts drama, whose embedding is closer to Shakespeare than to Tolstoï novels. The version of Hamlet presented here is fully commented, and thus is closer to analytical and philosophical works of Nietzsche and Plutarch as shown on the figure.
We also represent the variance learnt by the model in the size of the author dot. Hugo, who wrote famous novels as well as poetry and dramas, has a greater variance than other authors. 

\subsection{Interpretability of the Representation Space}

As we use the $L_2$ distance between document representations and stylistic feature vectors, each of the 300 embedding axes correspond to one given stylistic feature. The soft contrastive loss allows to ensures the $L_2$ constraint (bringing document embedding and stylistic features vectors closer) while being more flexible than a simple regression loss. When experimenting with the latter, the task showed up to be too hard and disadvantageous regarding both authorship attribution scores and writing style loss.

On Figure \ref{fig:correlation}, we show the Pearson correlation score between the $i^{th}$ stylistic feature and the corresponding embedding axis. These correlation values are always maximum for each feature regarding every other embedding coordinate. To further illustrate the interpretability of the embedding space, Figure \ref{fig:correlation_selection} shows a selection of 4 stylistic features, the representation value of the matching coordinate for each author. The representation space learnt by \modelname\ is interpretable in terms of writing style. In the context of a multidisciplinary project, involving several searchers in literature and linguistic this is a significant added value.

\subsection{Results on the Authorship Attribution Task}
Results on the authorship attribution task for IMDb62 and Blog Authorship Corpus are presented respectively in Table \ref{imdb62-results} against state-of-the-art solutions (not necessarily embedding models). On both datasets, our model ranks in top 4, outperforming recent competitors while authorship attribution is not its main task. Our model is outpaced by Syntax CNN \cite{zhang-etal-2018-syntax}, DBert-ft \cite{hay-etal-2020-representation} and BertAA \cite{fabien-etal-2020-bertaa}, two variants of BERT fine tuned on the authorship attribution task. As shown by \cite{fabien-etal-2020-bertaa}, BERT and DistilBERT are really tailored for balanced datasets with short texts such as IMDB62 and Blog Authorship Corpus. The DBert-ft model splits every document in 512 chunks during training, building an even bigger corpus with important improvement, but it is hardly reproducible with our feature loss. BertAA feeds encoded documents from a finetuned BERT together with a set of stylistic features and of most frequent bi-grams and tri-grams to a Logistic Regression. It clearly allows to better perform on Blog Authorship Corpus as this dataset is a mix of several genres and styles, compared to IMDB62 concerning only movie reviews. This confirms our use of stylistic features. Syntax CNN encodes each sentence of a document separately with its syntax. Unfortunately, this model was hardly reproducible and cannot be tested in feature regression using intermediate representation. For \modelname\, lower values of $\alpha$ allow to reach the best accuracy in authorship attribution on these datasets. Additional information bring by stylistic features benefit to the authorship attribution when texts are longer.

\subsection{Ablation Study and Effect of $\alpha$}
\begin{figure}
\centering
\includegraphics[width=0.8\columnwidth]{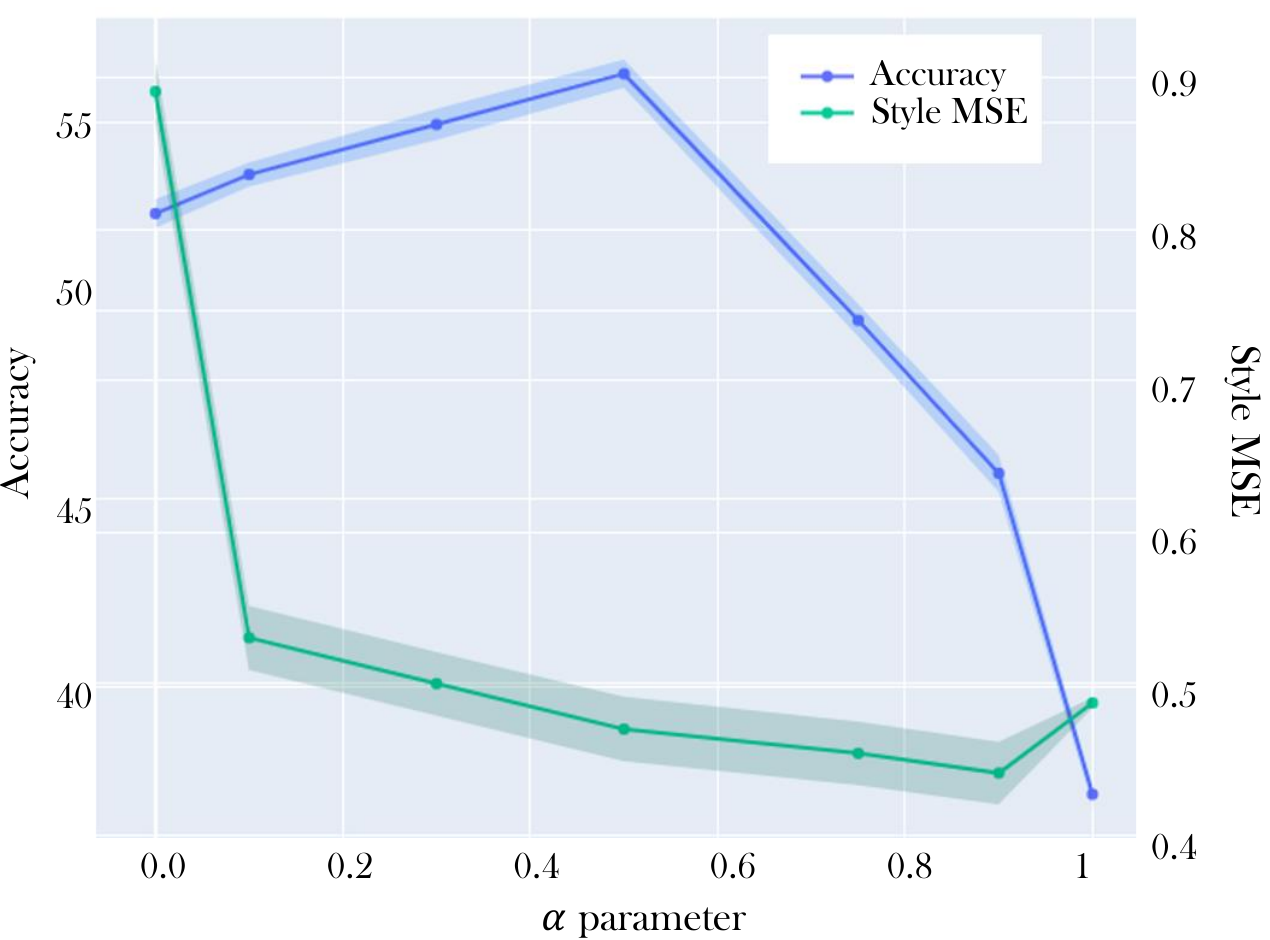}
\caption{\label{lrap_vs_mse}
\textbf{Effect of $\mathbf{\alpha}$.} We plot the evolution of the style evaluation metric (average MSE score) and of the accuracy with the $\alpha$ parameter for R-PGD}
\end{figure}

We here compare our model to no-VIB and without feature loss. Both variations underperform on both tasks. First, the VIB paradigm offers more versatility than fixed document and author representation which
is key to grasp a complex notion such as writing style. Then, the feature loss brings additional information for authorship prediction, as shown by BertAA, which use it to improve BERT classification results. Here, our framework enable to use it directly for document and author embeddings.
On Figure \ref{lrap_vs_mse}, we evaluate the influence of $\alpha$ which balances the importance given to author loss and feature loss on both feature regression and authorship attribution. Adding just a few stylistic features information ($\alpha=0.1$) allows to improve the precision of our model in authorship attribution. It forces the model to extract discriminant stylistic information from the input. Surprisingly the same phenomenon appears when shutting down the author loss ($\alpha = 1$). It creates a deterioration of the style score as authors tend to use a consistent writing style among their documents. Thus gathering a writer with its documents representation also helps to capture its writing habits. (\cite{hay-etal-2020-representation} call it the \textit{``Intra-author consistency''}).

\section{Conclusion}

In this article, we presented \modelname, a new author and document embedding method which leverages stylistic features. It has several advantages compared to existing works: it easily integrates any pretrained text encoder, it allows to compare authors and documents of any length (e.g., for authorship attribution), build an interpretable representation space by incorporating widely used stylistic features in computational linguistic. It is also able to infer representations for unseen documents at the opposite of most prior approaches. We demonstrated that \modelname\ outperforms existing embedding baselines in stylistic feature prediction, often by a large margin, while staying competitive in authorship attribution. 

In further experiments, we will incorporate modern text encoders, such as LLaMA \cite{touvron2023llama}. They are much more difficult to adapt to this task, but as most recent Large Language Model are trained in an autoregressive way, they might have the expressive power needed to grasp stylistic aspects of authors productions. 

\small{\subsubsection*{Acknowledgements}
This work was granted access to the HPC/AI resources of IDRIS under the allocation 2023-AD011012369R2 made by GENCI.
The author(s) acknowledge(s) the support of the French Agence Nationale de la Recherche (ANR), under grant ANR-19-CE38-0007 (project LIFRANUM).}

\bibliography{ecai}
\end{document}